\documentclass[letterpaper, 10 pt, conference]{ieeeconf}
\IEEEoverridecommandlockouts                              % This command is only needed if 
\usepackage{graphicx}
\graphicspath{{figures/}}
\usepackage{microtype}
\usepackage{cite}

\usepackage[absolute,showboxes]{textpos}
\TPMargin{5pt}

\newcommand{\copyrightstatement}{
    \begin{textblock}{0.84}(0.08,0.93)    % tweak here: {box width}(leftposition, rightposition)
         \noindent
         \footnotesize
         \copyright 2021 IEEE. Personal use of this material is permitted. Permission from IEEE must be obtained for all other uses, in any current or future media, including reprinting/republishing this material for advertising or promotional purposes, creating new collective works, for resale or redistribution to servers or lists, or reuse of any copyrighted component of this work in other works. doi: 10.1109/IROS51168.2021.9636838.
    \end{textblock}
}

\newcommand{\ignore}[1]{%
}

\title{\LARGE \bf Mobile Teleoperation: Feasibility of Wireless Wearable Sensing of the Operator's Arm Motion}
\author{Guanhao Fu$^1$, Ehsan Azimi$^2$, Peter Kazanzides$^2$
  \thanks{$^1$ Dept. of Mechanical Engineering, $^2$ Dept. of Computer Science,
          Johns Hopkins University, Baltimore USA, {\tt\footnotesize pkaz@jhu.edu}}%
}

\begin{document}

\copyrightstatement

\maketitle

\begin{abstract}
Teleoperation platforms often require the user to be situated at a fixed location to both visualize and control the movement of the robot and thus do not provide the operator with much mobility. One example is in existing robotic surgery solutions that require the surgeons to be away from the patient, attached to consoles where their heads must be fixed and their arms can only move in a limited space. This creates a barrier between physicians and patients that does not exist in normal surgery. To address this issue, we propose a mobile telesurgery solution where the surgeons are no longer mechanically limited to control consoles and are able to teleoperate the robots from the patient bedside, using their arms equipped with wireless sensors and viewing the endoscope video via optical see-through head-mounted displays (HMDs). We evaluate the feasibility and efficiency of our user interaction method compared to a standard surgical robotic manipulator via two tasks with different levels of required dexterity. The results indicate that with sufficient training our proposed platform can attain similar efficiency while providing added mobility for the operator.
\end{abstract}

\section{Introduction}

\begin{figure*}[t]
    \centering
    \hfill
    \includegraphics[width=0.45\linewidth]{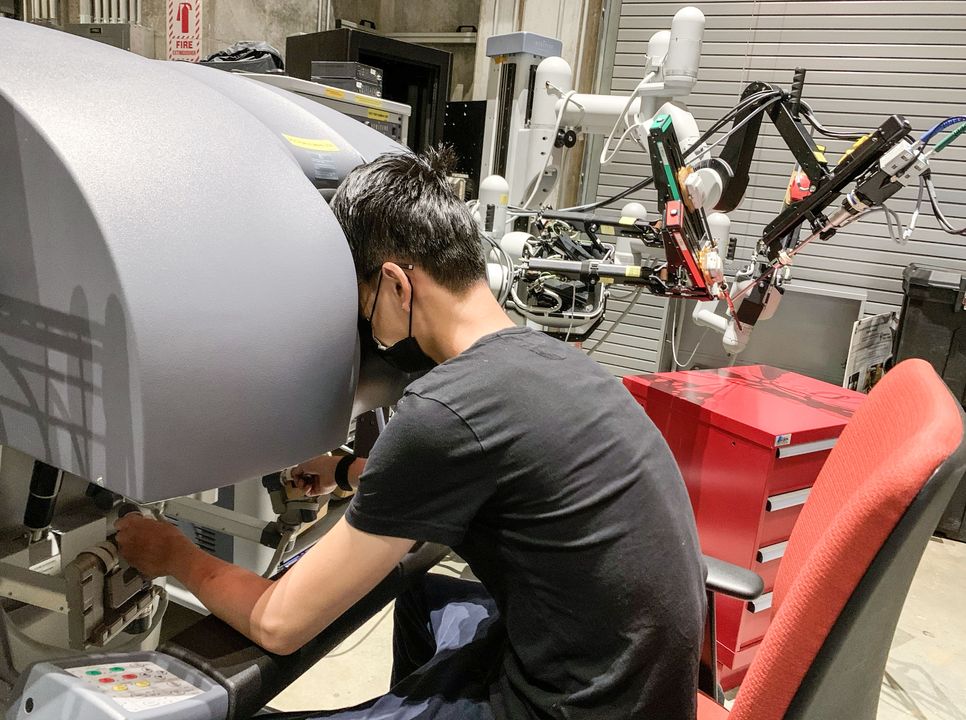}
    \hfill
    \includegraphics[width=0.44\linewidth]{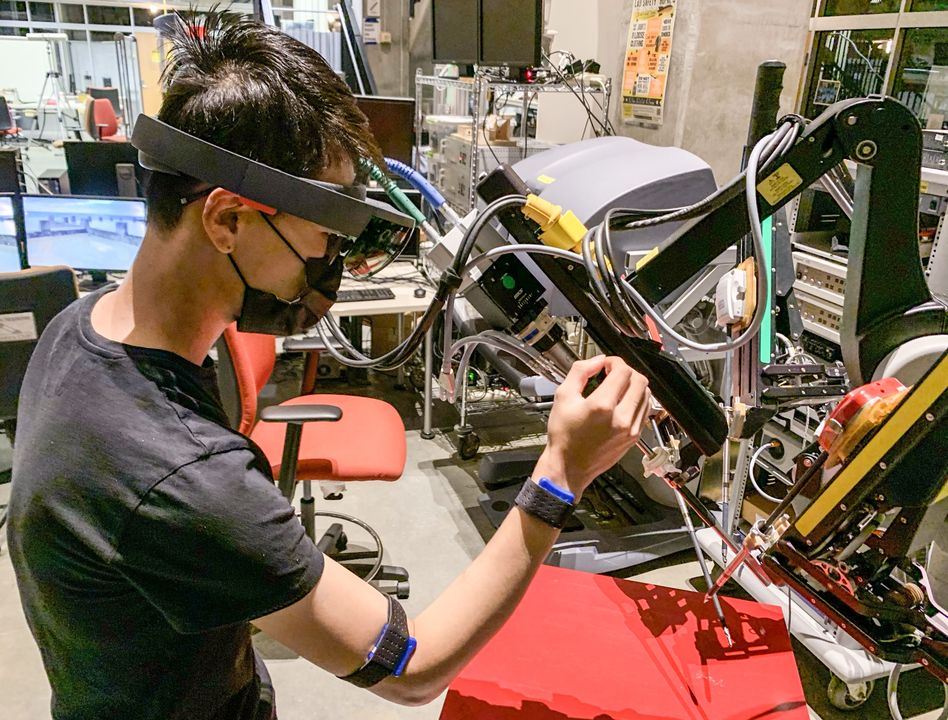}
    \hfill
    \caption{\textbf{Left:} Current teleoperation system where user is fixed to console and situated away from the robot. \textbf{Right:} Proposed mobile teleoperation system where user wears stereoscopic see-through HMD, situated next to the robot and operates the robot using a wearable motion capture system.}
    \label{fig:Holo_Rob}
\end{figure*}

Teleoperation is used in many applications where human presence must be extended to otherwise inaccessible areas, such as remote or dangerous locations (e.g., undersea, in space, or geographically distant) or small spaces (e.g., inside the patient's body during minimally-invasive surgery). At a minimum, these systems require a master console for the human operator to view camera images of the remote environment and then send motion commands to the remote robot(s).
%A typical implementation consists of one or more computer screens for viewing remote images and one or more input devices for sending motion commands. The input device may be as simple as a keyboard/mouse or game controller, or more complex such as one or more haptic devices. 
An example of a more complex master console is provided by the da Vinci Surgical System (Intuitive Surgical, Sunnyvale CA) \cite{Guthart2000}, shown in Fig.~\ref{fig:Holo_Rob}-left, where the primary surgeon sits at the master console, views stereo video on two displays (one for each eye), and controls the position of the remote robotic instruments using two Master Tool Manipulators (MTMs), which are 7 degree-of-freedom (dof) motorized mechanical linkages with an encoded passive gripper.
%
%Currently, when using the da Vinci surgical robot, the surgeon sits at the master console and teleoperates the patient-side robotic instruments while visualizing the anatomy via the stereo endoscope.
In this scenario, the surgeon is not scrubbed (not sterile) and is located away from the patient, thereby requiring 
%another surgeon or
an assistant to be present at the patient bedside.

\ignore{
\begin{figure}[b]
    \centering
    \includegraphics[width=\linewidth]{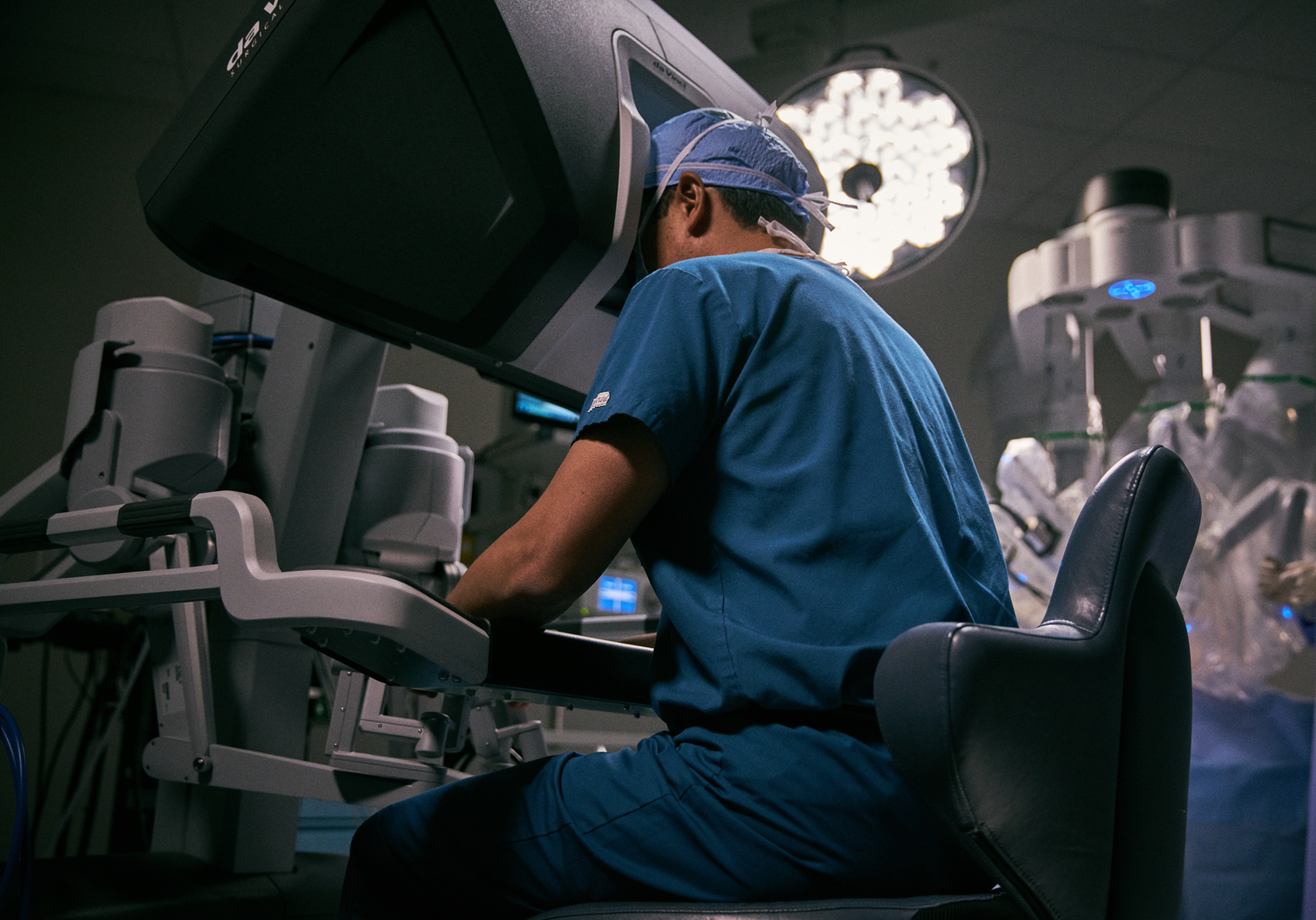}
    \caption{Surgeon at master console of da Vinci Surgical System, with patient-side robots in background. Image (C) 2020 Intuitive Surgical, Inc.}
    \label{fig:daVinci}
\end{figure}
}

The da Vinci provides one motivating example for a portable master console, as shown in Fig.~\ref{fig:Holo_Rob}-right,
because it would enable the surgeon to sit at the bedside, scrubbed,
and be able to directly interact with the patient when necessary. Clinically, this would enable
``solo'' surgeries, where the primary surgeon is able to perform the complete procedure, without a bedside assistant. More generally, however, the existence of a portable console would provide advantages in other telerobotic scenarios, such as disaster response, bomb disposal, or remote assistance for the disabled or elderly.

%The recent advances in 
A head-mounted display (HMD) provides an obvious component for a portable master console, as it enables visualization of the remote environment. Modern HMDs also include sophisticated sensing, such as multiple cameras and depth sensors, and recent advances in hand tracking have enabled natural gesture-based interactions. However, at present, it is not clear that HMD-based hand tracking would be sufficiently robust for high-precision tasks, such as surgery, and whether the requirement to keep the hands in the field-of-view of the HMD sensors would be comfortable for the operator. For example, a recent study used the HoloLens sensors to detect fine movements of the hands and fingers as well as touch interaction~\cite{MRTouch_Rob}, but this method requires a line of sight to the hand and can introduce uncertainty.  %but reported a 5\,mm measurement error, which is unsuitable for high-precision control.

We therefore propose the use of body-mounted inertial measurement units (IMUs) to provide the motion input for controlling the remote robots. We note that teleoperation interfaces may require the operator to specify more than the 6 dof positions of the remote robots. For example, in the da Vinci example cited above, it is also necessary to actuate the robotic instrument (e.g., open or close the jaws). We do not address this requirement here but, depending on the application, these additional actions could be implemented in a number of ways, including voice command, foot pedal, myoelectric sensing, or via a hand-held wireless device. At the same time, in order to have real mobility, a teleoperation system should be co-located anytime anywhere with the surgeon and therefore a wearable form factor is devised. Other lightweight setups, such as Phantom Omni, do not provide the mobility or portability of a wearable as they still must be fixed to a base and are merely a smaller MTM.

% PK: Following is from a proposal that was focused on da Vinci. In this paper, we may wish to play down the da Vinci since the proposed system can have broader use.
%At the same time, it also facilitates collaboration since multiple surgeons can wear HMDs and have access to a virtual console, thus providing the same benefits of the dual master console available on more recent da Vinci systems (but easily extended to more than two surgeons).

%We also contend that the virtual console can enable a more efficient, and adaptable, user interface, especially since it is not constrained by hardware implementations.
% Finally, the virtual console can increase the availability of training because the trainee would not require a physical console. Instead, training would be performed on the same system used in surgery, which is an HMD and a body-mounted input device.

%The goal is to provide a master console that is as portable as possible, requiring a minimal amount of hardware, and enabling the surgeon to remain sterile for the procedure.

%There are two distinct challenges: (1) tracking motion of the hand, which determines the position and orientation of the instrument, and (2) actuating the instrument, such as closing the jaws of a needle driver.

Related work includes a hand-held device, developed by Steidle et al. \cite{Steidle2016}, that
was tracked by fusion of optical and inertial sensing. The advantage of this approach
is that it can provide measurements to drive both the position and actuation of the instrument,
especially since the hand-held device can be designed with an appropriate interface mechanism.
One significant disadvantage, however, is the weight of the
hand-held device, which can cause fatigue during prolonged use. In addition, the
tracking technology may not be sufficiently reliable for critical tasks such as surgery. Their
system fused optical and inertial sensing, which can suffer from obstruction of the line-of-sight and
intermittent erroneous readings (e.g., due to computer vision failures), which can only be compensated
by the inertial sensing for brief periods of time. Electromagnetic tracking
would avoid the line-of-sight constraint, but be susceptible to electromagnetic field distortion.
Another disadvantage is that the operator would need to put down the hand-held device to perform
other tasks, and then have to pick it up again to continue teleoperation.

In addition, an IMU-based teleoperation system was introduced in \cite{Noccaro2017}; however, that work focused on using human arm motion to resolve the kinematic redundancy of a 7 degree-of-freedom robot and their experiments did not demonstrate the precise operation that would be required for surgical tasks.

This paper first presents our system design in Section \ref{sec:system}, including the kinematics of the IMU sensing system and a method for calibrating the link lengths of the human arm. Section \ref{sec:experiments} then describes the experimental setup to evaluate calibration accuracy, followed by experiments that compare the proposed IMU-based system to a da Vinci Master Tool Manipulator (MTM) to teleoperate virtual objects in two simulated training tasks. The results of those evaluations and comparative studies are presented in Section \ref{sec:results}, followed by the discussion and conclusions in Section \ref{sec:conclusions}.

\section{System Description}
\label{sec:system}

We propose an alternative master console to teleoperate robotic devices that uses IMUs as the input devices and Microsoft HoloLens as the visualization device. Ultimately, we envision a 6 IMU system that has 3 IMUs for each arm of the user,  which provides 6 dof Cartesian space control of robotic devices. Additionally, robotic instrument actuation will be addressed by a hand-held gripper device.
% that adds 1 dof to the system. 

%Currently, this paper aims to provide a fundamental structure for the proposed system. In particular, 
In this paper, we use 2 IMUs to achieve Cartesian space control of a virtual object, where the position and orientation of the virtual object reflect the user’s wrist pose. In the next iteration, a third IMU will be included so that our system can account for the 2 dof at the human wrist, which are wrist abduction/adduction, and wrist flexion/extension.

Based on our prior experience with inertial sensing \cite{Ren2012a, Ren2012b, He2015},
we realize that it is challenging to obtain accurate orientation. First, all inertial sensors (accelerometer, gyroscope, magnetometer) are subject to drift. Second, while a magnetometer (digital compass) provides an absolute measurement of heading, it is subject to magnetic field distortion. The alternative is to integrate the gyroscope reading, which is inaccurate due to drift (bias). In this particular application, however, the operator is controlling the position of a remote robotic end-effector with real-time visual feedback of the end-effector position. We hypothesize that this human visual-feedback loop would be tolerant of measurement drift because it would compensate for the induced error. The goal of our experiments is to provide evidence to support this hypothesis.

\begin{figure}[!t]
    \centering
    \includegraphics[width=0.95\linewidth]{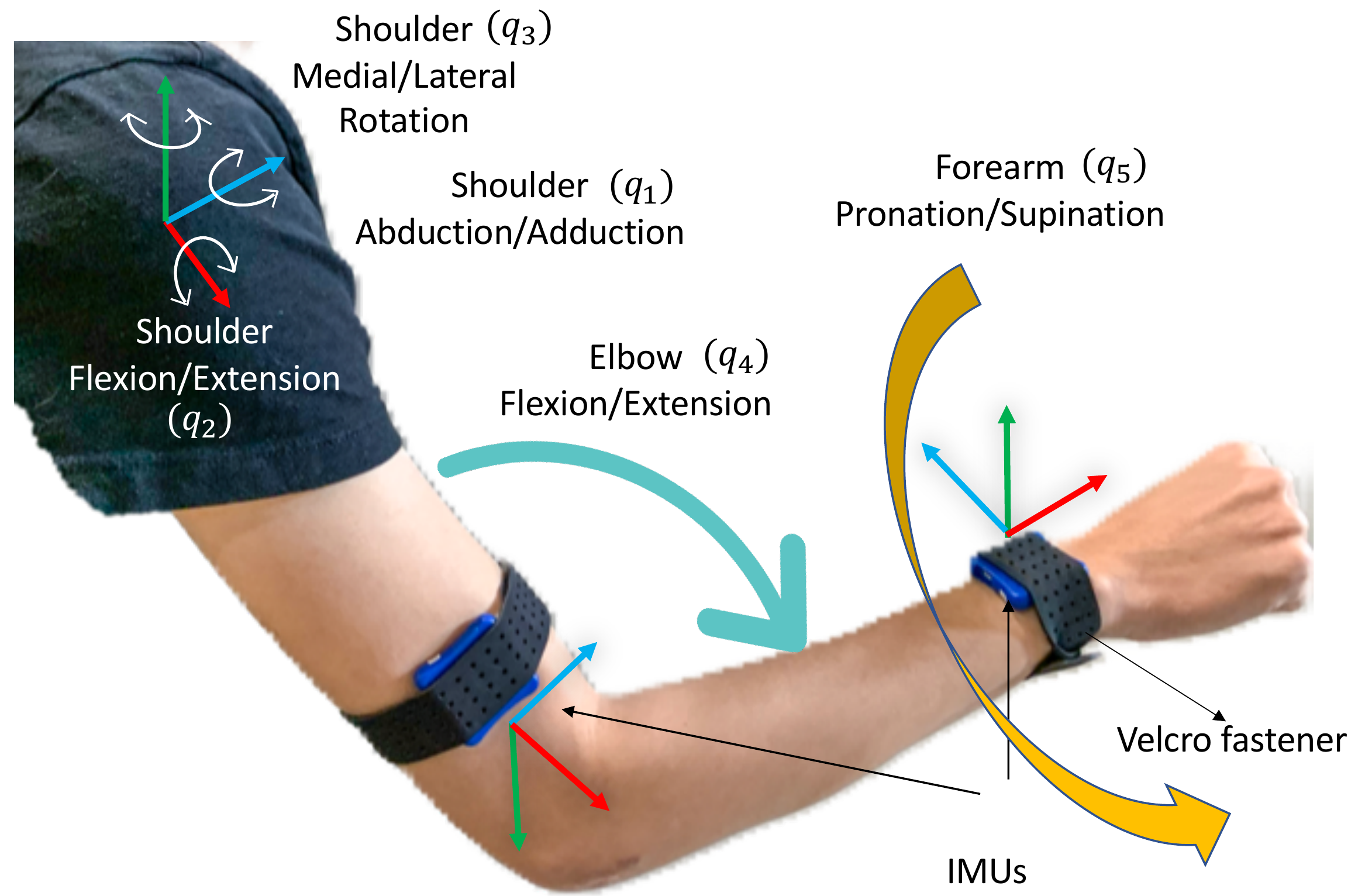}
    \caption{Experimental Setup: Two wireless IMUs are attached to the
upper arm and the forearm}
    \label{fig:exp_arm}
\end{figure}

\subsection{IMU System Kinematic Model}
The two IMUs (LPMS-B2, LP-RESEARCH Inc., Japan) are strapped onto the user's forearm and upper-arm (see Fig.~\ref{fig:exp_arm}), and the orientation outputs after fusing the raw data from accelerometer and gyroscope are used to obtain the end-effector position and orientation. Since there are 2 IMUs, this system is capable of capturing 5 dof based on a simplified human arm kinematic model \cite{Bertomeu-Motos2018}, which includes shoulder abduction/adduction ($q_{1}$), shoulder flexion/extension ($q_{2}$), shoulder medial/lateral rotation ($q_{3}$), elbow flexion/extension ($q_{4}$), and forearm pronation/supination ($q_{5}$).
% as referred to in Fig. \ref{fig:exp_arm}.

\begin{figure}[tbh]
    \centering
    \includegraphics[width=0.85\linewidth]{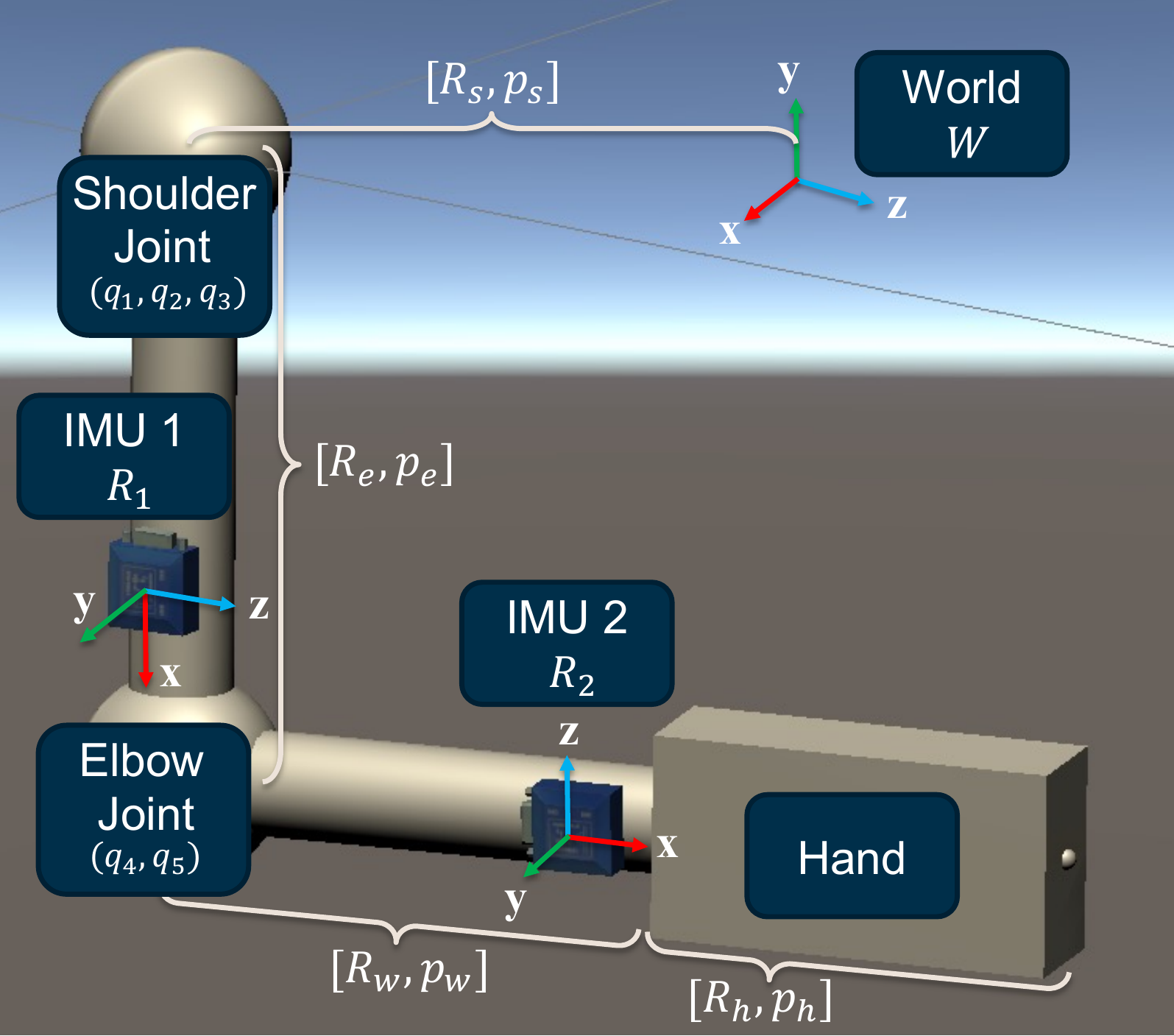}
    \caption{Simplified ball joint representation of human arm}
    \label{fig:balljoint}
\end{figure}

In the proposed system, however, joint angles are not directly measured by the IMU and the remote robot kinematics do not match the human arm kinematics, so only arm pose in Cartesian space is required. Thus, a simplified ball joint representation of the human arm is introduced instead, as shown in Fig. \ref{fig:balljoint}. The shoulder joint has 3 dof: $q_{1}$, $q_{2}$, and $q_{3}$, and the elbow joint has 2 dof: $q_{4}$ and $q_{5}$. 

The position and orientation of the user's wrist, $T_{wp} = F[R_{wp}, p_{wp}]$ w.r.t. the world coordinate frame is given by: 
\begin{equation} \label{eq:1}
T_{wp} = F[R_{s},p_{s}]\cdot F[R_{e},p_{e}]\cdot F[R_{w},p_{w}]
\end{equation}
 
where $R_{s} = R_{1}$, $R_{e} = R_{1}^{-1} \cdot R_{2}$, and $R_{w}$ is an identity rotation
% PK: a more complete system would define rotations between the IMUs and links, but at least we state our assumptions below.
(see Fig.~\ref{fig:balljoint} for frame definitions). Here, $p_{s} = [0, 0, 0]^T$, since we assume the center of the shoulder joint is aligned with the world coordinate's origin. $R_{1}$ and $R_{2}$ are the orientation of IMU 1 and IMU 2 w.r.t. the world coordinate frame.  We also assume that the IMUs are aligned with the axes of the upper-arm and forearm which requires careful alignment in our experiments; if necessary, existing calibration methods, such as \cite{Mueller2017}, can be used to compensate for any misalignment. 

Here, the user's upper-arm and forearm lengths, $l_{u}$ and $l_{f}$, are needed to compute the user's wrist pose:
\begin{equation} \label{eq:2}
p_{e} = [l_{u}, 0, 0]^T \qquad p_{w} = [l_{f}, 0, 0]^T
\end{equation}

The values of $l_{u}$ and $l_{f}$ can be obtained by the calibration procedure described in the next section or by physical measurement of the operator's arm link lengths. 

\subsection{User's Arm Length Calibration}

\begin{figure}[t]
    \centering
    \includegraphics[width=0.81\linewidth]{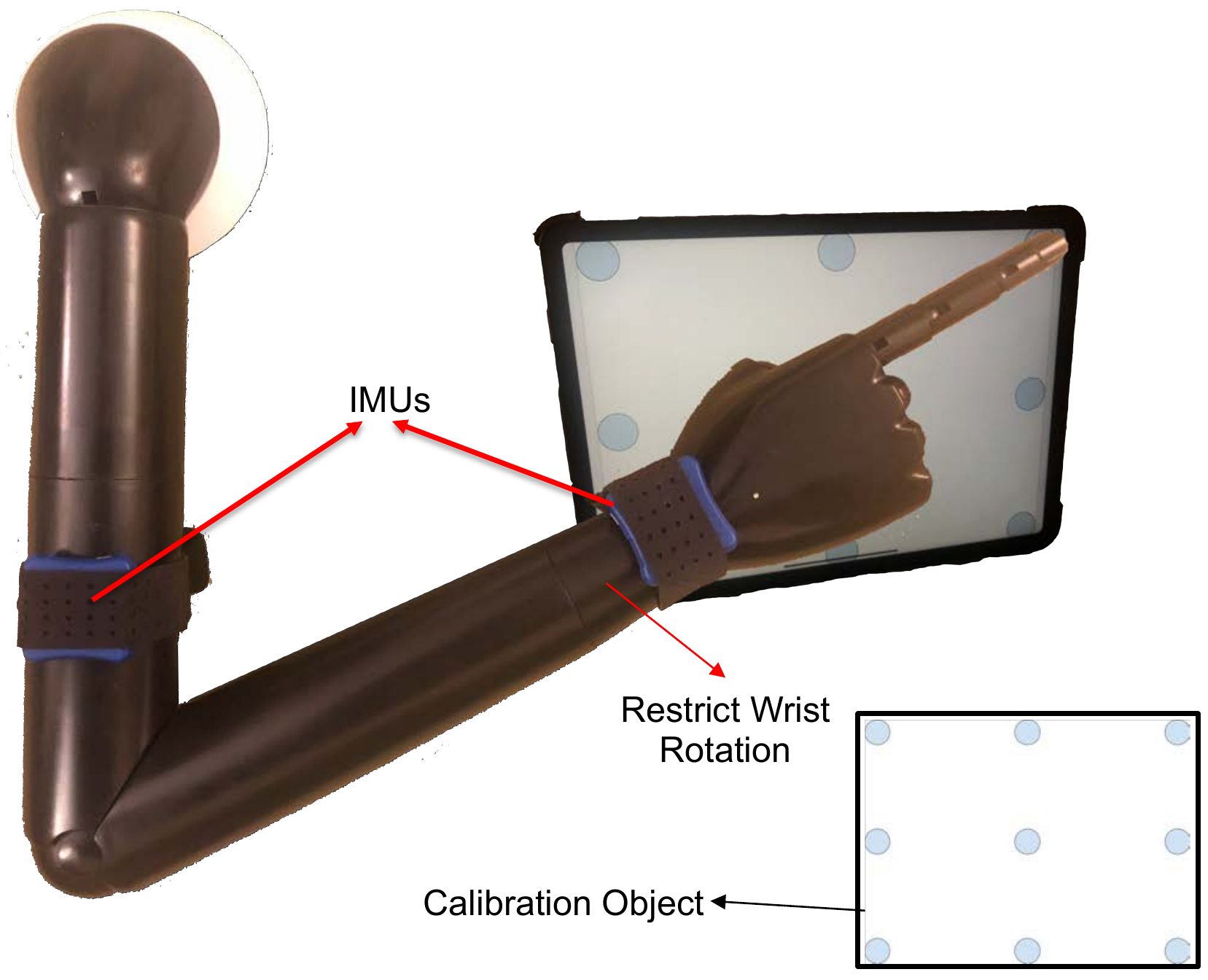}
    \caption{Calibration setup using mannequin arm }
    \label{fig:mannequin}
    \vspace{-6pt} 
\end{figure}

%Intuitively, different users have different arm lengths. 
%In order for our kinematic model to provide an accurate representation of the user's wrist pose, we need to obtain the user's individual arm link lengths. In particular, 
The developed calibration method requires users to touch at least 4 of the 9 different points shown in Fig. \ref{fig:mannequin}, which can be printed or shown on any flat surface, as long as the physical distances between the points can be accurately measured. Users should not rotate their hip or torso during the calibration procedure because the current 2 IMU system assumes that the user's shoulder joint is fixed in position and orientation. 
More importantly, the user needs to touch different points with fully extended index finger, and without rotating the wrist, then record the orientation of the 2 IMUs $R_{1}$ and $R_{2}$ at each calibration point on the object. 
The reason that we chose to use the user's extended index finger is because it is a relatively intuitive way for users to touch designated points on an object. We extend Eq. (\ref{eq:1}) to include the hand and fingertip as follows:

\begin{equation} \label{eq:4}
T_{ft} = F[R_{s},p_{s}]\cdot F[R_{e},p_{e}]\cdot F[R_{w},p_{w}]\cdot F[R_{h},p_{h}]
\end{equation}
\begin{equation} \label{eq:5}
p_{h} = [l_{h}, 0, 0]^T
\end{equation}

$T_{ft} = F[R_{ft}, p_{ft}]$ is the position and orientation of the user's fingertip w.r.t. the world frame and $p_{h}$ is the user's hand length measured from the wrist. 
% DF: changed wording here, please check
$R_{h}$ is an identity rotation because wrist rotation is restricted during the calibration and we use a measured value of 0.2\,m for $l_h$. 

When the user touches a calibration point $i$ on the calibration object, the fingertip position $p_{ft}(i)$ is recorded. Then, we compute the distance between where the user's fingertip touches points $i$ and $j$, where $i \neq j$, and denote it as the distance from the forward kinematics: $d_{FK}$.
\begin{equation} \label{eq:6}
d_{FK}(i,j) = \left | p_{ft}(i) - p_{ft}(j) \right | , i \neq j, i,j \in [1,2,...,9]
\end{equation}

Similarly, we denote the true physical distance between the same two calibration points as:
\begin{equation} \label{eq:7}
d_{tr}(i,j) = \left | p_{tr}(i) - p_{tr}(j) \right | , i \neq j, i,j \in [1,2,...,9]
\end{equation}

We use the Matlab \texttt{fmincon} optimizer to find the $l_{u}$ and $l_{f}$ that minimizes: 
\begin{equation} \label{eq:8}
\sum_{i,j} \left \| d_{FK}(i,j) - d_{tr}(i,j) \right \|
\end{equation}

\section{Experiments}

The following sections present the experiments performed to evaluate the calibration accuracy and teleoperation performance of our IMU-based system.
%to the Master Tool Manipulator (MTM) from the da Vinci Research Kit (dVRK), an open source research platform based on the first-generation da Vinci surgical robot \cite{Kazanzides2014}.

\begin{figure*}[t]
    \centering
    \hfill
    \includegraphics[width=0.39\linewidth]{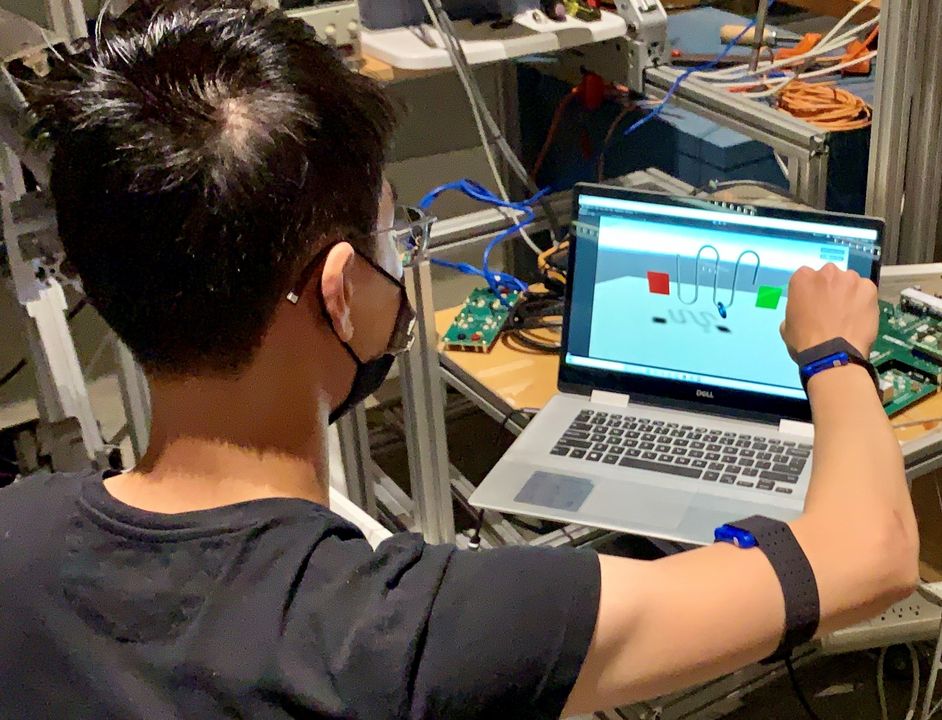}
    \hfill
    \includegraphics[width=0.535\linewidth]{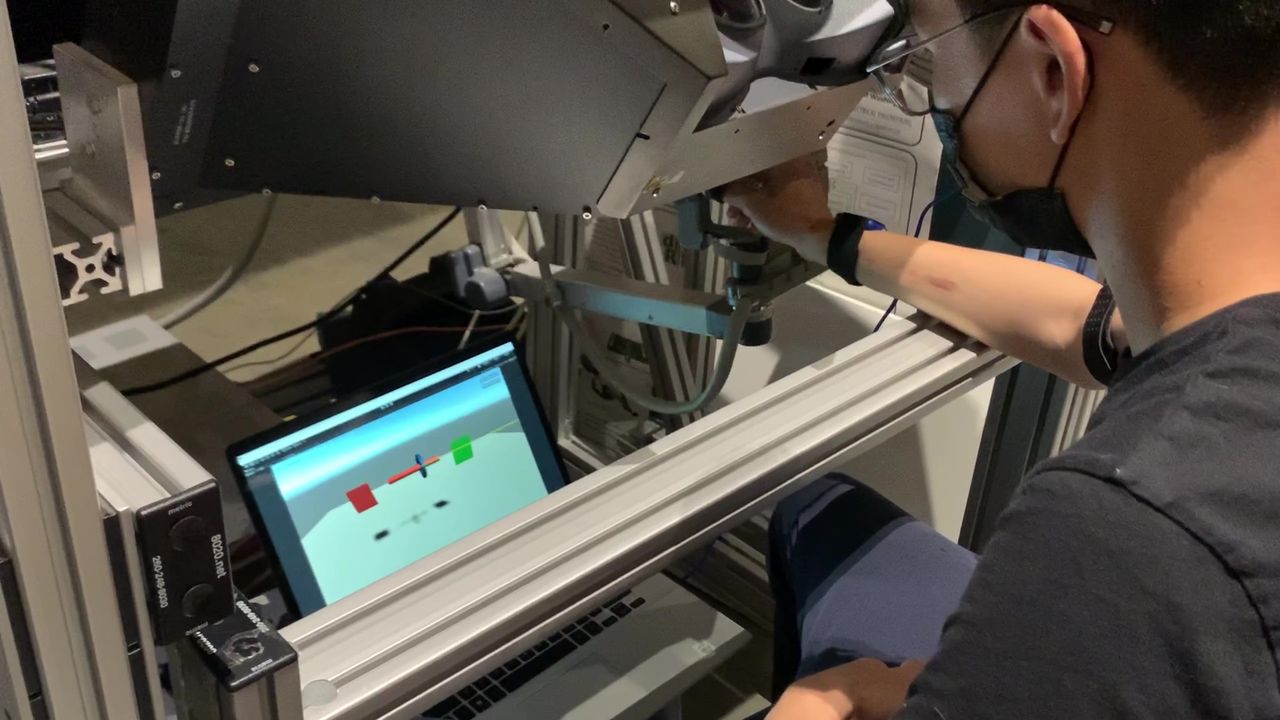}
    \hfill
    \caption{Experimental setup: User performing the curved wire task using proposed system (Left) and the straight wire task using da Vinci MTM (Right)}
    \label{fig:exp_mtm_imu}
\end{figure*}

\label{sec:experiments}
\subsection{Calibration Accuracy}

To evaluate our calibration method, we mounted the IMUs on a mannequin arm and manually moved the arm to touch 4 of the 9 calibration posts shown in Fig. \ref{fig:mannequin}. We used a mannequin arm because accurate measurement on a human arm is challenging due to skin artifact introducing unwanted IMU movement~\cite{Mueller2017, Lopez-Nava2016, El-Gohary2012}. It is also difficult to obtain accurate ground-truth measurements of human arm lengths.

\subsection{Teleoperation Performance}

% Here you can describe the Unity scene you created and the experiments.
After calibrating the user's arm lengths, we evaluate the performance of our system compared to the MTM from the da Vinci Research Kit (dVRK), an open source research platform based on the first-generation da Vinci surgical robot \cite{Kazanzides2014}, as seen in Fig. \ref{fig:exp_mtm_imu}.

To capture the performance of our proposed system, we designed two tasks with different levels of required dexterity. We measured both accuracy in position and orientation as well as task completion time.

\subsubsection{Visualization Setup}
\begin{figure}[tbh]
    \centering
    \includegraphics[width=0.88\linewidth]{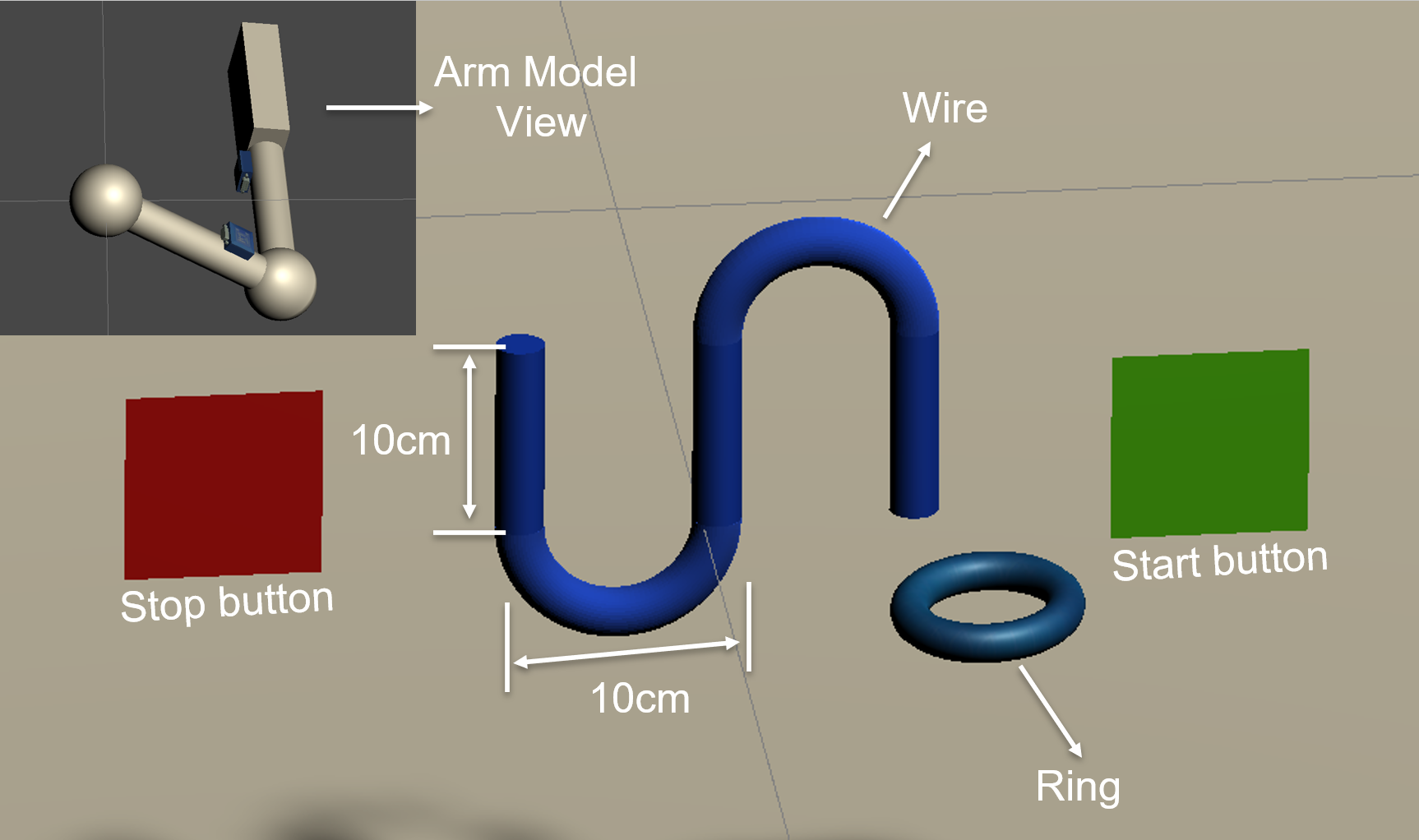}
    \caption{Unity scene setup: Steady hand task with S-shaped wire}
    \label{fig:unity_view}
    \vspace{-6pt} 
\end{figure}

We designed the tasks around a classic steady-hand game that can often be found in surgical robotics training curricula \cite{Mariani2018}. The tasks consist of two main objects, a ring and a wire, and the user is required to move the ring along the wire without any collision (see accompanying video). The tasks are visualized in Unity3D, with intuitive start and stop buttons that automatically record time stamped data to evaluate position and orientation accuracy. As shown in Fig. \ref{fig:unity_view}, there is also a simultaneous view of the above mentioned simplified ball joint representation of the user's arm configuration. It is important to note that the visualization is in 2D, with a slight pan angle for improved depth perception. 

\begin{figure}[t]
    \centering
    \includegraphics[width=0.88\linewidth]{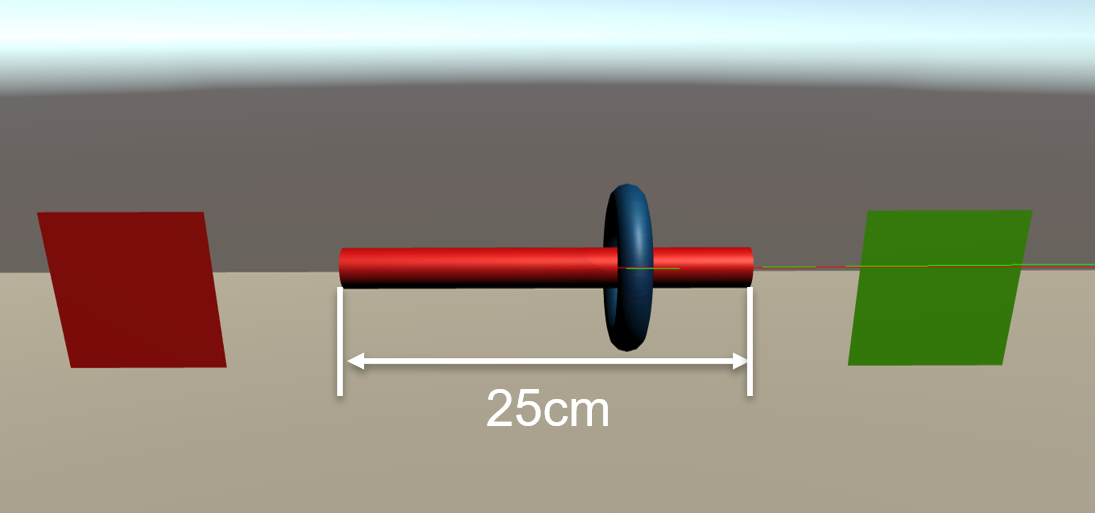}
    \caption{Unity scene setup: Steady hand task with straight wire }
    \label{fig:task_easy}
    \vspace{-6pt} 
\end{figure}

In Fig. \ref{fig:task_easy}, the wire is straight and is horizontally oriented. In Fig. \ref{fig:unity_view}, the wire is S-shaped, which requires a higher level of dexterity, thus serving as a good evaluation of our system's ability to perform more complex tasks.
In both task setups, the wires have 25\,mm diameter, and the rings have 60\,mm inner diameter (ID) and 100\,mm outer diameter (OD), which makes the collision threshold 17.5\,mm. The wire shape dimensions are shown in their respective figures. 

\subsubsection{Input Devices Setup}
% \begin{figure}[t]
%     \centering
%     \includegraphics[width=\linewidth]{IMU_wire.jpg}
%     \caption{Experimental Setup: User performing the curved wire task using proposed system}
%     \label{fig:exp_imu}
% \end{figure}

% \begin{figure}[t]
%     \centering
%     \includegraphics[width=\linewidth]{MTM_line.jpg}
%     \caption{Experimental Setup: User performing the straight wire task using da Vinci MTM}
%     \label{fig:exp_mtm}
% \end{figure}

We interfaced the IMUs with Unity3D using the Unity plugin developed by LP-RESEARCH Inc. In addition, we streamed the Cartesian position and orientation of the MTM through User Datagram Protocol (UDP) and then converted from a right handed coordinate system to the left handed coordinate system used by Unity3D. We let both input devices directly control the ring object in Unity, without any medium such as grippers used by the da Vinci surgical robot. 

\section{Results}
\label{sec:results}
\subsection{Calibration Results}
We performed 10 trials using our calibration method, where we manually moved the mannequin arm to touch 4 of the 9 calibration posts shown in Fig. \ref{fig:mannequin}. Table ~\ref{tab:calibration} shows the errors in the estimated arm lengths, including the average and standard deviation.
% of error in arm lengths are reported in table~\ref{tab:calibration}.

\begin{table}
\vspace{3pt}  % Fix margin when at top of page
\centering
\caption{Calibration results: link length errors as percentages of the true lengths}
\begin{tabular}{|l|r|r|}
 \hline
& Upper arm & Forearm \\
 \hline
Trial 1 &3.1\% &  11.3\% \\
 \hline
Trial 2 &0.9\% &  4.7\% \\
 \hline
Trial 3 &24.6\% &  7.7\% \\
 \hline
Trial 4 &14.4\% &  15.5\% \\
 \hline
Trial 5 &21.7\% &  6.0\% \\
 \hline
Trial 6 &1.6\% &  11.5\% \\
 \hline
Trial 7 &0.6\% &  4.4\% \\
 \hline
Trial 8 &2.1\% &  0.3\% \\
 \hline
Trial 9 &9.1\% &  0.2\% \\
 \hline
Trial 10 &18.7\% &  16.4\% \\
 \hline
Mean     &9.7\% &  7.8\% \\
 \hline
Std Dev. &9.4\% &  5.7\% \\
 \hline
\end{tabular}
\label{tab:calibration}
\vspace{-6pt}
\end{table}

\subsection{Teleoperation results}
Two users performed each task three times.
%, as our University safety protocols do not allow that). 
User1 was familiar with the IMU system, but had no prior experience with the MTM, while User2 was a novice with the IMU but had some experience with the MTM.
We evaluated the position and orientation accuracy \cite{Enayati2018} of both our system and the dVRK while performing the two tasks described above.
The position accuracy, at each point on the motion trajectory, is defined as the distance between $c_{ring}$ and $c_{wire}$: 
\begin{equation} \label{eq:9}
distance(c_{wire}, c_{ring})
\end{equation}
where $c_{ring}$ is the ring center, and $c_{wire}$ is the point on the wire center-line that is closest to $c_{ring}$. 
The orientation accuracy is defined as the angle between the ring orientation $ \vec{v}_{ring}$ and the wire tangent line $\vec{v}_{wire}$:
\begin{equation} \label{eq:10}
\alpha = cos^{-1} \left ( \frac{\vec{v}_{wire}\cdot \vec{v}_{ring}}{\left \| \vec{v}_{wire} \right \|\cdot \left \| \vec{v}_{ring} \right \|} \right )
\end{equation}

\begin{figure}[b]
    \centering
    \includegraphics[width=0.94\linewidth]{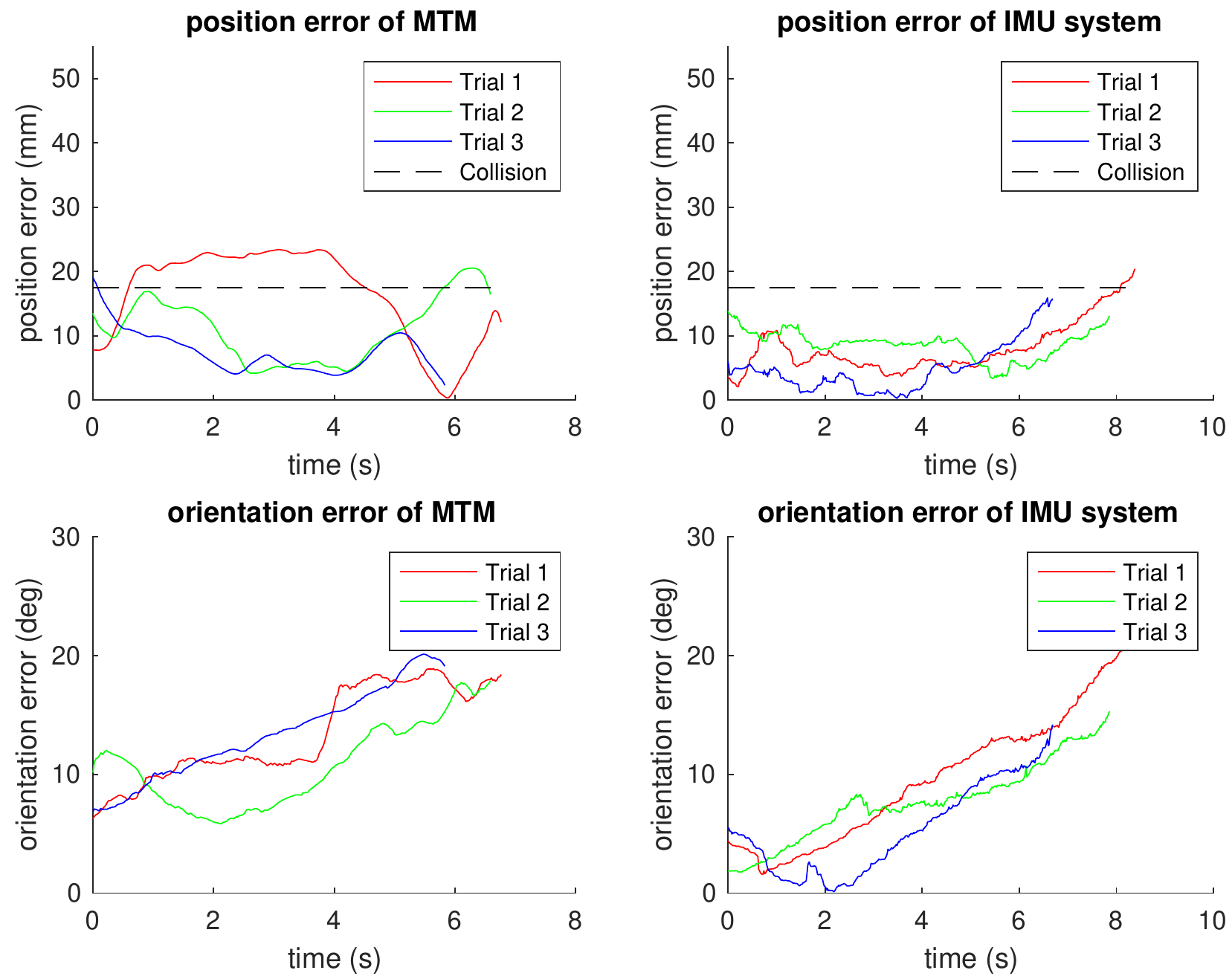}
    \caption{Result of straight wire teleoperation task for User1}
    \label{fig:res_st_users}
\end{figure}

\begin{figure}[t]
    \centering
    \includegraphics[width=0.91\linewidth]{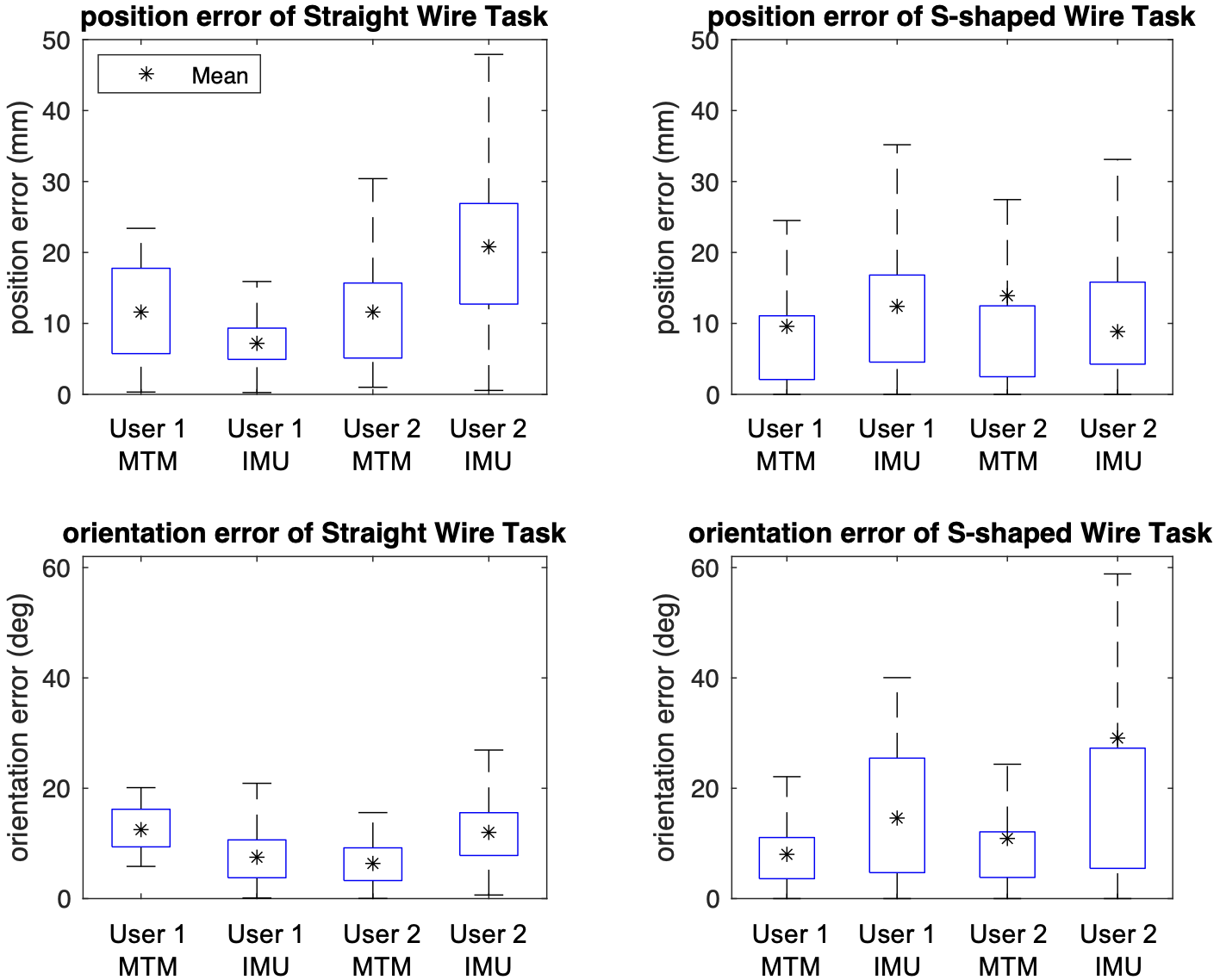}
    \caption{Accuracy comparison of both users performing two tasks}
    \label{fig:box}
\end{figure}

The variation for position and orientation errors with time are presented only for the straight line task for User1 in Fig.~\ref{fig:res_st_users}, due to space limitations. The results for both tasks performed by both users are summarized in Fig.~\ref{fig:box}. 
Task completion times for the straight wire and S-shaped wire are provided in Table~\ref{tab:task-time}.
The result shows that it took longer for User2 to complete the task.
Table~\ref{tab:taskPos} and Table~\ref{tab:taskOri} depict the mean position and orientation error for the tasks.
%, User2 shows the higher position and orientation accuracy than User1 while using the MTM. This indicates that task completion time increases while errors decrease. 

\begin{table}
\centering
\caption{Completion times for tasks, in seconds}
\begin{tabular}{|l|rrrr|rrrr|}
 \hline
 & \multicolumn{4}{|c}{User1 Trials} & \multicolumn{4}{|c|}{User2 Trials} \\
 & 1 & 2 & 3 & Mean & 1 & 2 & 3 & Mean \\
 \hline
 \multicolumn{9}{|c|}{\emph{Straight Wire Task}} \\
 \hline
 MTM & 6.8 & 6.6 & 5.8 & 6.4 & 9.6 & 11.1 & 8.4 & 9.7 \\
 \hline
 IMU & 8.4 & 7.9 & 6.7 & 7.6 & 7.2 & 9.6 & 10.8 & 9.2 \\
 \hline
%\end{tabular}
%\label{tab:Stime}
%\vspace{-6pt}
%\end{table}

%\begin{table}
%\centering
%\caption{Completion times for S-shaped Wire task, in seconds}
%\begin{tabular}{|l|rrrr|rrrr|}
 %\hline
 %& \multicolumn{4}{|c}{User1 Trials} & \multicolumn{4}{|c|}{User2 Trials} \\
 %& 1 & 2 & 3 & Mean & 1 & 2 & 3 & Mean \\
 \multicolumn{9}{|c|}{\emph{S-shaped Wire Task}} \\
 \hline
 MTM & 19.7 & 18.1 & 16.2 & 18.0  & 24.6 & 22.0 & 21.1 & 22.5 \\
 \hline
 IMU & 26.6 & 26.5 & 21.0 & 24.7 & 25.2 & 28.4 & 23.8 & 25.8 \\
 \hline
\end{tabular}
%\label{tab:SStime}
\label{tab:task-time}
\vspace{-6pt}
\end{table}

\begin{table}
\centering
\caption{Mean position errors for tasks, in millimeters}
\begin{tabular}{|l|rrrr|rrrr|}
 \hline
 & \multicolumn{4}{|c}{User1 Trials} & \multicolumn{4}{|c|}{User2 Trials} \\
 & 1 & 2 & 3 & Mean & 1 & 2 & 3 & Mean \\
 \hline
 \multicolumn{9}{|c|}{\emph{Straight Wire Task}} \\
 \hline
 MTM & 16.3 & 11.0 & 7.6 & 11.6 & 11.0 &  7.0 & 16.8 & 11.6 \\
 \hline
 IMU & 8.1 & 8.7 & 4.8 & 7.2 & 27.7 & 14.9 & 19.7 & 20.8 \\
 \hline
%\end{tabular}
%\label{tab:SPos}
%\end{table}

%\begin{table}
%\centering
%\caption{Mean position errors for S-shaped Wire task, in millimeters}
%\begin{tabular}{|l|rrrr|rrrr|}
% \hline
% & \multicolumn{4}{|c}{User1 Trials} & \multicolumn{4}{|c|}{User2 Trials} \\
% & 1 & 2 & 3 & Mean & 1 & 2 & 3 & Mean \\
 \multicolumn{9}{|c|}{\emph{S-shaped Wire Task}} \\
 \hline
 MTM & 9.7 &  10.8  &  8.2  &  9.6 &  16.1 &  12.8 &  12.8  & 13.9 \\
 \hline
 IMU & 9.4 &  10.2 &  17.6  & 12.4  &  9.1  &  9.9 &  7.5  & 8.8 \\
 \hline
\end{tabular}
\label{tab:taskPos}
\vspace{-6pt}
\end{table}

\begin{table}
\centering
\caption{Mean orientation errors for tasks, in degrees}
\begin{tabular}{|l|rrrr|rrrr|}
 \hline
 & \multicolumn{4}{|c}{User1 Trials} & \multicolumn{4}{|c|}{User2 Trials} \\
 & 1 & 2 & 3 & Mean & 1 & 2 & 3 & Mean \\
 \hline
 \multicolumn{9}{|c|}{\emph{Straight Wire Task}} \\
  \hline
 MTM & 13.4 & 10.7 & 13.3 & 12.5  &  6.5  &  5.9  &  6.7  &  6.4 \\
 \hline
 IMU & 9.6 &   7.7 &   5.2 &  7.5 &  12.7  & 11.4  & 11.8 & 12.0 \\
 \hline
%\end{tabular}
%\label{tab:SOri}
%\end{table}

%\begin{table}
%\centering
%\caption{Mean orientation errors for S-shaped Wire task, in degrees}
%\begin{tabular}{|l|rrrr|rrrr|}
% \hline
% & \multicolumn{4}{|c}{User1 Trials} & \multicolumn{4}{|c|}{User2 Trials} \\
% & 1 & 2 & 3 & Mean & 1 & 2 & 3 & Mean \\
 \multicolumn{9}{|c|}{\emph{S-shaped Wire Task}} \\
 \hline
 MTM &  6.5 &   6.9 &  10.7  &  8.0 &  10.6 &  10.9 &  11.3 &  10.9\\
 \hline
 IMU &  13.7 &  14.8  & 15.3 &  14.6 &  30.8 &  28.3 &  28.1 & 29.1 \\
 \hline
\end{tabular}
\label{tab:taskOri}
\vspace{-6pt}
\end{table}

\begin{table}
\vspace{4pt}
\centering
\caption{Non-collision percentage for straight wire task}
\begin{tabular}{|l|rr|rr|}
 \hline
 & User1 & User2\\
 \hline
 MTM &  74.5\% &  85.8\% \\
 \hline
 IMU &  98.7\% &  47.4\% \\
 \hline
\end{tabular}
\label{tab:percentage}
\end{table}

% PK: Some of this should be in Discussion section
When comparing the performance for the same task and the same user across different input devices, as represented in Fig. \ref{fig:box} and Tables~\ref{tab:taskPos} and~\ref{tab:taskOri}, we observe that User1, who is more familiar with the system, performs similarly with both systems. %we see that the IMU system has slightly higher mean errors than the MTM, except for User1 using the IMU system while doing the straight wire task. 
This suggests that the proposed IMU system could have comparable position accuracy to the MTM, once the learning curve is completed. For the orientation error, especially while the users were doing the S-shaped wire task, the IMU system has larger errors, which is due to the limited dexterity in the system: our IMU system only captures 5 dof, whereas the MTM has 7 dof. 
Dashed lines in Fig.~\ref{fig:res_st_users} represent the threshold where the ring collides with the wire and Table~\ref{tab:percentage} shows the ``success rate'' measure, which is the percentage of time that the ring did not collide with the wire.

\section{Discussion and Conclusions}
\label{sec:conclusions}

Table~\ref{tab:calibration} shows that, on average, the calibration method can estimate the link lengths within 10\% of the true values. However, this table also shows significant variation in the result, even though each trial was performed with the calibration object in approximately the same location. Better repeatability could possibly be obtained by using more than 4 calibration points, at the cost of a longer calibration time. From the results of the teleoperation experiments, however, we can conclude that the current calibration accuracy is sufficient for the evaluated tasks because performance was comparable to a conventional input device, with differences that could be attributed to user familiarity.

Table~\ref{tab:percentage} indicates that despite the inherent challenges with IMUs, a trained user can perform the task with a high success rate. For a novice user, performance with the MTM was generally better than with the IMU. Of course, participants using IMU input were limited by the range of motion of human arms and by the boundaries of their workspace. One possible way to address this issue is to introduce a clutch input so that when users reach the workspace limit, they can clutch and reorient themselves in a neutral location to have room to continue the task. 

Limitations of this feasibility study include the small number of participants (two), limited trials per task (three), and differences in participant experience and possible effect of a learning curve. Furthermore, this study did not quantify the IMU sensor drift, so it is uncertain how much visual compensation was required.

While the IMU-based system provided comparable performance to the MTM in terms of allowing the user to specify motion of the remote object, unlike the mechanically-grounded MTM, it cannot provide haptic feedback.
One possible solution is to add vibrotactile actuators, for example,
on the IMUs.
Other possible solutions include sensory substitution (e.g., graphical overlays \cite{Akinbiyi2006} or audio cues \cite{Balicki2010, Cutler2013} to indicate measured force) or force feedback to some other part of the body, such as the forearm or wrist \cite{Ng2007}.

In summary, this study evaluated the feasibility of mobile teleoperation using wireless IMUs mounted on the user's arm by comparing the efficiency of the system against a standard mechanical input device used for robotic surgery. The results show that our proposed solution is a trade-off that adds considerable mobility while introducing some inaccuracy, though the inaccuracy may be mitigated by user training and stereoscopic visualization. This study provides evidence that although an IMU-based input device is subject to drift, it can effectively be used in teleoperation scenarios where the operator is closing the control loop based on visual feedback. We are currently integrating the IMU-based system with a HoloLens HMD to provide a complete teleoperation solution. Stereoscopic visualization provided by the HMD can potentially mitigate the errors caused by a lack of sufficient depth perception in our current visualization. We also intend to add a clutch function to address the workspace limitation and to conduct a controlled multi-user study.

\section*{Acknowledgments}
%Anton Deguet assisted with the software and system integration of the experimental platform.
Anton Deguet assisted with the system integration.

%\clearpage

% Can use IEEEtranS for sorted bibliography
\bibliographystyle{IEEEtran}
\bibliography{references}

\begin{thebibliography}{10}
\providecommand{\url}[1]{#1}
\csname url@rmstyle\endcsname
\providecommand{\newblock}{\relax}
\providecommand{\bibinfo}[2]{#2}
\providecommand\BIBentrySTDinterwordspacing{\spaceskip=0pt\relax}
\providecommand\BIBentryALTinterwordstretchfactor{4}
\providecommand\BIBentryALTinterwordspacing{\spaceskip=\fontdimen2\font plus
\BIBentryALTinterwordstretchfactor\fontdimen3\font minus
  \fontdimen4\font\relax}
\providecommand\BIBforeignlanguage[2]{{%
\expandafter\ifx\csname l@#1\endcsname\relax
\typeout{** WARNING: IEEEtran.bst: No hyphenation pattern has been}%
\typeout{** loaded for the language `#1'. Using the pattern for}%
\typeout{** the default language instead.}%
\else
\language=\csname l@#1\endcsname
\fi
#2}}

\bibitem{Guthart2000}
G.~Guthart and J.~K. Salisbury~Jr, ``The {Intuitive}\texttrademark~ telesurgery
  system: Overview and application.'' in \emph{IEEE Intl. Conf. on Robotics and
  Automation (ICRA)}, 2000, pp. 618--621.

\bibitem{MRTouch_Rob}
R.~Xiao, J.~Schwarz, N.~Throm, A.~D. Wilson, and H.~Benko, ``{MRTouch}: Adding
  touch input to head-mounted mixed reality,'' \emph{IEEE Trans. on Vis. \&
  Comp. Graphics}, vol.~24, no.~4, pp. 1653--1660, 2018.

\bibitem{Steidle2016}
F.~Steidle, A.~Tobergte, and A.~Albu-Sch\"{a}ffer, ``Optical-inertial tracking
  of an input device for real-time robot control,'' in \emph{IEEE Intl. Conf.
  on Robotics and Automation (ICRA)}, May 2016, pp. 742--749.

\bibitem{Noccaro2017}
A.~Noccaro, F.~Cordella, L.~Zollo, G.~{Di Pino}, E.~Guglielmelli, and
  D.~Formica, ``{A teleoperated control approach for anthropomorphic
  manipulator using magneto-inertial sensors},'' in \emph{IEEE Intl. Symp. on
  Robot and Human Interactive Comm. (RO-MAN)}, 2017, pp. 156--161.

\bibitem{Ren2012a}
H.~Ren, D.~Rank, M.~Merdes, J.~Stallkamp, and P.~Kazanzides, ``Multisensor data
  fusion in an integrated tracking system for endoscopic surgery,'' \emph{IEEE
  Trans. on Info. Tech. in Biomed.}, vol.~16, no.~1, pp. 106--111, Jan 2012.

\bibitem{Ren2012b}
H.~Ren and P.~Kazanzides, ``Investigation of attitude tracking using an
  integrated inertial and magnetic navigation system for hand-held surgical
  instruments,'' \emph{IEEE/ASME Trans. on Mechatronics}, vol.~17, no.~2, pp.
  210--217, Apr 2012.

\bibitem{He2015}
C.~He, P.~Kazanzides, H.~T. Sen, S.~Kim, and Y.~Liu, ``An inertial and optical
  sensor fusion approach for six degree-of-freedom pose estimation,''
  \emph{Sensors}, vol.~15, no.~7, pp. 16\,448--16\,465, Jul 2015.

\bibitem{Bertomeu-Motos2018}
A.~Bertomeu-Motos, A.~Blanco, F.~J. Badesa, J.~A. Barios, L.~Zollo, and
  N.~Garcia-Aracil, ``Human arm joints reconstruction algorithm in
  rehabilitation therapies assisted by end-effector robotic devices,'' \emph{J.
  of NeuroEngin. \& Rehab.}, vol.~15, no.~1, pp. 1--11, 2018.

\bibitem{Mueller2017}
P.~Mueller, M.-A. Begin, T.~Schauer, and T.~Seel, ``Alignment-free,
  self-calibrating elbow angles measurement using inertial sensors,''
  \emph{IEEE J. of Biomedical \& Health Info.}, vol.~21, no.~2, pp. 312--319,
  2017.

\bibitem{Lopez-Nava2016}
I.~H. Lopez-Nava and M.~M. Angelica, ``{Wearable Inertial Sensors for Human
  Motion Analysis: A review},'' \emph{IEEE Sensors Journal}, vol.~16, no.~22,
  pp. 7821--7834, 2016.

\bibitem{El-Gohary2012}
M.~El-Gohary and J.~McNames, ``{Shoulder and elbow joint angle tracking with
  inertial sensors},'' \emph{IEEE Transactions on Biomedical Engineering},
  vol.~59, no.~9, pp. 2635--2641, 2012.

\bibitem{Kazanzides2014}
P.~Kazanzides, Z.~Chen, A.~Deguet, G.~S. Fischer, R.~H. Taylor, and S.~P.
  DiMaio, ``An open-source research kit for the {da
  Vinci}\textregistered~surgical system,'' in \emph{IEEE Intl. Conf. on
  Robotics and Automation (ICRA)}, Hong Kong, China, Jun 2014, pp. 6434--6439.

\bibitem{Mariani2018}
A.~Mariani, E.~Pellegrini, N.~Enayati, P.~Kazanzides, M.~Vidotto, and E.~{De
  Momi}, ``{Design and Evaluation of a Performance-based Adaptive Curriculum
  for Robotic Surgical Training: A Pilot Study},'' in \emph{IEEE Engin. in
  Medicine and Biology Conf. (EMBC)}, 2018, pp. 2162--2165.

\bibitem{Enayati2018}
N.~Enayati, A.~M. Okamura, A.~Mariani, E.~Pellegrini, M.~M. Coad, G.~Ferrigno,
  and E.~{De Momi}, ``Robotic assistance-as-needed for enhanced visuomotor
  learning in surgical robotics training: An experimental study,'' in
  \emph{IEEE Intl. Conf. on Robotics and Automation (ICRA)}, 2018, pp.
  6631--6636.

\bibitem{Akinbiyi2006}
T.~Akinbiyi, C.~E. Reiley, S.~Saha, D.~Burschka, C.~J. Hasser, D.~D. Yuh, and
  A.~M. Okamura, ``Dynamic augmented reality for sensory substitution in
  robot-assisted surgical systems,'' in \emph{IEEE Engin. in Medicine and
  Biology Conf. (EMBC)}, 2006, pp. 567--570.

\bibitem{Balicki2010}
M.~Balicki, A.~Uneri, I.~Iordachita, J.~Handa, P.~Gehlbach, and R.~Taylor,
  ``Micro-force sensing in robot assisted membrane peeling for vitreoretinal
  surgery,'' in \emph{Medical Image Computing and Computer-Assisted
  Intervention (MICCAI)}, 2010, pp. 303--310.

\bibitem{Cutler2013}
N.~Cutler, M.~Balicki, M.~Finkelstein, J.~Wang, P.~Gehlbach, J.~McGready,
  I.~Iordachita, R.~Taylor, and J.~T. Handa, ``Auditory force feedback
  substitution improves surgical precision during simulated ophthalmic
  surgery,'' \emph{Investigative Ophthalmology \& Visual Science}, vol.~54,
  no.~2, pp. 1316--1324, Feb. 2013.

\bibitem{Ng2007}
G.~Ng, P.~Barralon, G.~Dumont, S.~K.~W. Schwarz, and J.~M. Ansermino,
  ``Optimizing the tactile display of physiological information: Vibro-tactile
  vs. electro-tactile stimulation, and forearm or wrist location,'' in
  \emph{IEEE Engin. in Med. and Bio. Conf. (EMBC)}, 2007, pp. 4202--4205.

\end{thebibliography}

\end{document}